\pdfoutput=1

\documentclass[11pt]{article}

\usepackage{acl}

\usepackage{times}
\usepackage{latexsym}

\usepackage[T1]{fontenc}

\usepackage[utf8]{inputenc}

\usepackage[nolist]{acronym} 
\usepackage{hyperref}
\usepackage{kotex}
\usepackage{booktabs}
\usepackage{array,multirow,graphicx}
\usepackage{float}
\usepackage{makecell}
\usepackage{soul}
\usepackage{color}
\usepackage{subcaption}
\usepackage{amsmath}
\usepackage{bbm}

\newcommand{\ins}[1]{\textcolor{blue}{#1}}
\newcommand{\inp}[1]{\textcolor{red}{#1}}

\title{\textsc{KoBEST}: Korean Balanced Evaluation of Significant Tasks}

\author{
Dohyeong Kim$^1$ ~~~
Myeongjun Jang$^2$ ~~~
Deuk Sin Kwon$^1$ ~~~
Eric Davis$^1$ ~~~
\\
$^1$Language Super Intelligence Labs, SK Telecom, South Korea \\
\{bruno, brian.k, eric.davis\}@sktair.com \\
$^2$Department of Computer Science, University of Oxford, UK \\
myeongjun.jang@cs.ox.ac.uk \\
}

\begin{document}
\maketitle
\begin{abstract}
A well-formulated benchmark plays a critical role in spurring advancements in the \ac{NLP} field, as it allows objective and precise evaluation of diverse models. As modern \acp{LM} have become more elaborate and sophisticated, more difficult benchmarks that require linguistic knowledge and reasoning have been proposed. However, most of these benchmarks only support English, and great effort is necessary to construct benchmarks for other low resource languages. To this end, we propose a new benchmark named \ac{KoBEST}, which consists of five Korean-language downstream tasks. Professional Korean linguists designed the tasks that require advanced Korean linguistic knowledge. Moreover, our data is purely annotated by humans and thoroughly reviewed to guarantee high data quality. We also provide baseline model and human performance results. Our dataset is available on the Huggingface~\footnote{\href{https://huggingface.co/datasets/skt/kobest_v1}{https://huggingface.co/datasets/skt/kobest\_v1}}.
\end{abstract}

\section{Introduction}\label{section:intro}
The \ac{NLP} field is now facing unprecedented rapid development. A major factor propelling the progress is the existence of unified benchmark datasets like \textsc{GLUE}~\cite{GLUE}, which are designed to assess models' language understanding capabilities. Such benchmark datasets, enabled modern \acp{PLM}, such as BERT \cite{BERT} and GPT-2 and GPT-3 \cite{GPT2, GPT3}, to be assessed in objective and multifaceted manners. The success of \textsc{GLUE} also lead to similar benchmark datasets in a variety of other languages.~\cite{CLUE, FLUE, indonlu, KLUE}.

However, many recent studies reveal that the outstanding performance of \acp{PLM} on such benchmark datasets seems plausible but not probable. These studies have found that datasets may contain many spurious artefacts, and the performance of \acp{PLM} is enhanced by  excessive usage of said artefacts~\cite{habernal2018argument, niven2019probing, mccoy2019right, benderkoller2020climbing}. Another line of work observed that many \acp{PLM}, which showed promising results in \textsc{GLUE}, fall short of expectations for more difficult tasks that require linguistic knowledge~\cite{bhatt2021case} or logical reasoning~\cite{logicNLI}. As a result, the importance of well-designed evaluation datasets with higher difficulty-level has been highlighted, and new datasets, such as \textsc{CheckList}~\cite{ribeiro2020beyond}, and \textsc{LogicNLI}~\cite{logicNLI}, have been proposed. Most of them only support specific languages like English, and it requires large efforts to build higher difficulty-level language evaluation suits for other low resource languages, however. 

When it comes to the Korean language, two benchmarks are widely used: Korean-NLI \& STS~\cite{ham2020kornli} and KLUE~\cite{KLUE}. The former is machine- and human-translated from English \ac{NLI} and \ac{STS} datasets, which hardly reflect the characteristics of the Korean language. The latter is a Korean version GLUE benchmark which supports eight tasks, such as \ac{NLI}, \ac{STS}, \ac{NER}, and \ac{RE}. Although these tasks are useful for assessing general language ability, it is difficult to ascertain whether a model is able to reason based on more complicated knowledge beyond text form (e.g., passage of time, meaning of text, causality). 
To this end, we aim to construct a new benchmark dataset in Korean named \ac{KoBEST}, which consists of five downstream tasks that require advanced knowledge of Korean. We carefully constructed the data based on the following design principles:

\begin{itemize}
    \item \textit{Human-driven data annotation}: Our data is purely annotated by humans to prevent incorrect and ambiguous data instances caused by automatic data annotation approach.
    
    \item \textit{Leveraging professional linguistic knowledge}: As a result of our collaboration with professional Korean linguists, we re able to collect grammatically correct data with rich vocabulary and expressions.
    
    \item \textit{Availability to public}: As a benchmark dataset, it is important to ensure public
    accessibility. We guarantee that our data is free to use and redistribute.
    
    \item \textit{High data quality}: Our data passed thorough reviews driven by both models and humans to deliver high quality data without superficial cues and heuristic artefacts.

    \item \textit{Avoiding AI ethical issues}: Our data does not contain any toxic content, social biases, or personal information.

\end{itemize}

Next, we evaluated widely used Korean \acp{PLM} on the \textsc{KoBEST} dataset. Specifically, we conducted fine-tuning, zero-shot, one-shot, and few-shot experiments. The experimental results can serve as a baseline for performance on \textsc{KoBEST}. Participants also provided human performance baselines for all of our tasks. Our results suggest that modern \acp{PLM} and a large-size \ac{GLM} are far from reaching human-level language ability.

\begin{table}[t!]
	\begin{center}
		\renewcommand{\arraystretch}{1.2}
		\footnotesize{
			\centering{\setlength\tabcolsep{0.8pt}
		\begin{tabular}{c|c|c|c|c|c}
		\toprule
		\multicolumn{1}{c|}{Tasks} & \multicolumn{1}{c|}{|Train|} &  \multicolumn{1}{c|}{|Dev|} &
		\multicolumn{1}{c|}{|Test|} & 
		\multicolumn{1}{c|}{Metrics} & 
		\multicolumn{1}{c}{Text Source}\\ \hline
		KB-BoolQ & 3.665 & 700 & 1,404 & F1 & Wikipedia  \\
		KB-COPA & 3,076 & 1,000 & 1,000 & F1 & N.A \\
		KB-WiC & 3,318 & 1,260 & 1,260 & F1 & Korean Dictionary \\
		KB-HellaSwag & 2,029 & 500 & 500 & F1 & \makecell[c]{Wikipedia, \\ YouTube}  \\
		KB-SentiNeg & 3.649 & 400 & 397 & F1 & Product reviews\\
		\bottomrule
		\end{tabular}}}
	\end{center}
	\caption{The number of data instances for each downstream task.}\label{table.data_size}
\end{table}

\section{KoBEST Downstream Tasks}\label{section:KoBEST}

\subsection{Overview}
The KoBEST benchmark consists of the following five downstream tasks:

\begin{enumerate}
    \item KoBEST-BoolQ (KB-BoolQ): identify whether a given question is true or false considering a paragraph.
    \item KoBEST-COPA (KB-COPA): select an alternative which is a cause/effect of a given premise.
    \item KoBEST-WiC (KB-WiC): identify whether the meaning of a target word is the same or different in two given contexts. 
    \item KoBEST-HellaSwag (KB-HellaSwag): select a correct sentence among four candidates that is likely to appear after a given context.
    \item KoBEST-SentiNeg (KB-SentiNeg): predict the polarity of a negated sentence.
\end{enumerate}

The number of training/development/test data points is illustrated in Table~\ref{table.data_size}.

\begin{table*}[t!]
	\begin{center}
		\renewcommand{\arraystretch}{1.2}
		\footnotesize{
			\centering{\setlength\tabcolsep{1.5pt}
		\begin{tabular}{c|l}
		\toprule
		\rotatebox[origin=c]{90}{\textbf{KB-BoolQ}} & \makecell[l]{\textbf{Paragraph}: 구한말, 통영 안뒤산 기슭 간창골에 김봉제 형제가 살았다. 김봉제는 관약국을 경영하며 부를 누렸는데, \\ 선비적 성품을 지녔던 형과 반대로 막냇동생 김봉룡은 성질이 포악했다. 어느 날 봉룡은 아내였던 숙정을 사모하던 \\ 나그네를 살해하였고, 숙정은 누명을 벗으려고 비상을 먹고 자살한다. (At the end of the Joseon Dynasty, the \\ Kim Bong-je brothers lived in Ganchanggol at the foot of Andui Mountain in Tongyeong. Kim Bong-je enjoyed wealth \\ while running a government office, but his younger brother Kim Bong-ryong had a violent temper, contrary to his brother \\ who had a scholarly character. One day, Bongryong killed a traveler who adored his wife Sukjeong, and Sukjeong commits \\ suicide to clear his name.) \\ \textbf{Question}: 봉룡은 숙정을 죽였는가? (Did Bongryong kill Sukjeong?) \quad \textbf{Answer}: False} \\ \hline
		
		\rotatebox[origin=c]{90}{\textbf{KB-COPA}} & \makecell[l]{\textbf{Premise}: 전쟁이 시작되었다. (The war had begun.) \quad \textbf{Question}: 결과 (Effect) \\ \textbf{Alternative 1}: 병사들이 집으로 돌아왔다. (Soldiers returned home.) \\ \textbf{Alternative 2}: 병사들이 전투에 파견되었다.  (Soldiers were sent to battle.) \\ \textbf{Correct Alternative}: 2} \\ \hline
		
		\rotatebox[origin=c]{90}{\textbf{KB-WiC}} & \makecell[l]{\textbf{Context 1}: 망가진 엔진은 \colorbox{pink}{수리}가 불가능하다. (It is impossible to repair a broken engine.)\\ \textbf{Context 2}: 이 배는 \colorbox{pink}{수리}에 들어간 지 일주일이 됐다. (The ship has been under repair for a week.) \\ \textbf{Target Word}: 수리 (repair) \quad \textbf{Answer}: True} \\ \hline
		
		\rotatebox[origin=c]{90}{\textbf{KB-HellaSwag}} & \makecell[l]{\textbf{Context}: 양궁 선수들이 경기장으로 입장한다. 관중들이 함성을 지르고 응원한다. 선수들이 상대팀과 악수하고 \\ 자리로 돌아온다. 코치가 전략을 설명하고 화이팅을 외친다. (Archery players enter the stadium. The crowd shouts  \\and cheers. The players shake hands with the opposing team and return to their seats. The coach explains the strategy \\ and shouts "Go for it.") \\ \textbf{Ending 1}: 활이 과녁에 적중했다. (The arrow hits the target.) \\ \textbf{Ending 2}: 선수가 심호흡을 하고 활을 쏜다. (The player takes a deep breath and shoots an arrow.) \\ \textbf{Ending 3}: 선수가 활을 들어 과녁을 조준한다. (The player raises his bow and aims at the target.) \\  \textbf{Ending 4}: 선수들이 각자 자리에 서서 활을 꺼낸다. (The players stand in their own positions and take out their bows.) \\ \textbf{Correct Ending}: 4} \\ \hline
		
		\rotatebox[origin=c]{90}{\textbf{KB-SentiNeg}} & \makecell[l]{\textbf{Sentence 1}: 뚜껑이 잘 열려요! (The lid opens well!) \quad \textbf{Label 1}: \ins{긍정 (Positive)} \\ \textbf{Sentence 2}: 뚜껑이 잘 안열려요! (The lid does not open well!) \quad \textbf{Label 2}: \inp{부정 (Negative)} \\} \\
		
		\bottomrule
		\end{tabular}}}
	\vspace*{-2ex}
	\caption{Examples of development set from the KoBEST tasks. The variables of each task are highlighted in bold. Text written in parenthesis is the English translated version of the original data points.}\label{table.examples}%
	\end{center}
	\vspace*{-4ex}
\end{table*}

\subsection{KoBEST-BoolQ}
\paragraph{Data/Task Description} We built the KB-BoolQ dataset by referencing \ac{BoolQ} task~\cite{boolq}. A data point consists of a paragraph, question, and label. The task aims to evaluate models' understanding of the paragraph by asking a yes/no question. An example is presented in Table~\ref{table.examples}.

We extracted paragraphs from Korean Wikipedia\footnote{\href{https://ko.wikipedia.org/wiki/}{https://ko.wikipedia.org/wiki/}}. To cover diverse topics, such as Science/Technology and Art/Culture, we defined keywords for each topic and selected documents containing enough information regarding the keyword. Next, we extracted paragraphs for each document and generated corresponding questions that could be answered as yes/no based on the paragraph.

\paragraph{Guidelines} Annotators were instructed to construct the KB-BoolQ dataset following the guidelines described below.
\begin{enumerate}
    \item Paragraphs should be evenly extracted from various domains and topics to minimise bias.
    
    \item Questions should be answered only with the information presented in the paragraph. We set this guide to exclude the impact of pre-trained commonsense knowledge for decision making. Also, annotators have different viewpoints regarding the boundary of commonsense knowledge, which can cause uneven task difficulty.
    
    \item Questions should be written in clear, unambiguous, easy-to-understand language. A true/false judgement should be obvious from a human perspective.
\end{enumerate}

\subsection{KoBEST-COPA}
\paragraph{Data/Task Description} We referenced \ac{COPA}~\cite{copa} to construct the KB-COPA dataset. The data has four variables: premise, two alternatives, and a question that asks a model to decide the cause or effect of the premise from the two alternatives. An example is available in Table~\ref{table.examples}.

\paragraph{Guidelines} We provided the following guidelines to annotators for generating data instances.
\begin{enumerate}
    \item The alternatives should belong to a similar area, e.g., states and actions. This rule is introduced to preclude systems from making decisions based on situational difference, not the meaning of alternatives.
    
    \item The alternatives should contain a keyword related to that of premise. For instance, in the example presented in Table~\ref{table.examples}, the keyword of the premise is ``전쟁 (war)'', and both alternatives contain the related same keyword ``병사 (soldier)''. We introduce this guideline to increase the task's difficulty by making the alternatives belong to the same category.
    
    \item All the premises and alternatives should be written concisely so that the content can be understood intuitively. Therefore, using proper noun, slang, and redundant expressions should be avoided.
    
    \item All sentences should be written in the past tense. In the Korean language, simple present can cause confusion because it has indication of tense and sometimes can imply  present progressive. On the other hand, the past tense is morphologically clear and is able to convey meaning without confusion.
    
    \item All sentences must include a subject. Although the subject is frequently omitted in Korean, it is difficult to infer the cause or effect without a subject because all the premises and alternatives are quite short. Therefore, even though such sentences are slightly unnatural in Korean, we guide annotators to insert a subject.
\end{enumerate}

\subsection{KoBEST-WiC}
\paragraph{Data/Task Description} KB-WiC is a task that determines whether a word has the same connotation in different contexts. We referenced \ac{WiC}~\cite{wic} when building the dataset. An instance is composed of a target homonym and two different contexts that contain the target word. Table~\ref{table.examples} provides an example for the KB-WiC task. Unlike the original \ac{WiC} dataset that includes various word forms, we only used words with the same form, so as to focus more on recognising a word's meaning without the distracton of variouus forms.

\paragraph{Guidelines} To construct the KB-WiC dataset, we instructed annotators to follow these guidelines.
\begin{enumerate}
    \item A target word should be listed in the National Institute of the Korean Language Basic Korean Dictionary\footnote{\href{https://stdict.korean.go.kr/main/main.do}{https://stdict.korean.go.kr/main/main.do}} or Korea University Korean Dictionary~\cite{hong2009korea}. We exclude words not registered in the dictionaries because they can cause ambiguous criteria for determining an answer. This is despite the fact that they are generally used in daily life.
    
    \item For generating a data point where an answer is \textit{False}, only a \textbf{homonym} should be used as a target word because a \textbf{polysemy} makes the task considerably more challenging, even for native speakers.
    
    \item The \ac{PoS} tag of a target word should be a noun, pronoun, numeral, or dependent noun\footnote{A noun that cannot be used without the help of other words in Korean.}. We introduce this guideline because the four \ac{PoS} tags have a fixed form and distinct meaning in Korean.
    
    \item The contexts should be extracted from example sentences in the dictionaries to make it possible to clearly understand the sense of a target word only using the given context.
\end{enumerate}

\subsection{KoBEST-HellaSwag}
\paragraph{Data/Task Description} This task evaluates whether a system can utilize passage of time and order to complete the last sentence in a series of sentences. We referenced the  HellaSwag dataset~\cite{hellaswag} to build our version but modified the task to consider specific characteristics of the Korean language.

The original HellaSwag benchmark was designed to ascertain whether a \ac{LM} can generate a plausible ending sentence given a relevant subject and context. In Korean, however, subjects are typically omitted. As a result, if the ending sentence is generated from the subject, the sentence becomes awkward and barely resembles a plausible Korean sentence. Evaluating such unnatural sentences is not in line with the purpose of \ac{KoBEST}, so we modify the task to predict the most plausible final sentence among four candidates. An example instance is available in Table~\ref{table.examples}.

\paragraph{Guidelines} We instruct annotators to build the data based on the following guidelines.
\begin{enumerate}
    \item The annotators should generate or modify free-text descriptions of YouTube videos and Wikepedia documents that progress with the passage of time.

    \item At least three sentences should be included in the context. A system should have as much context as possible to generate a plausible ending sentence.
    
    \item All the candidate-ending sentences should be thematically related to the context. The answer should only be able to be found by analysing the passage of time among the sentences, not via the topic or keywords. This guideline is introduced to prevent low task difficulty by generating alternative endings that simply contradict the correct ending.
\end{enumerate}

\subsection{KoBEST-SentiNeg}
\paragraph{Data/Task Description} Many studies have revealed that \acp{PLM} lack understanding of negation expressions~\cite{hossain2020analysis, ettinger2020bert, kassner2020negated, hosseini2021understanding}. Inspired by the \textit{Negation} capability test of \citet{ribeiro2020beyond}, we designed a similar but enhanced task by utilizing negation to create sentences opposite in meaning. Specifically, we created a two-class sentiment analysis task by generating product reviews based on real product reviews available on e-commerce websites.~\footnote{The real product reviews are only used as references and our data is newly generated by human annotators.} We then used the \textit{training} and \textit{development} sets to train a sentiment classification model. Next, we extracted candidates from training data where the polarity switched when transformed into a sentence with the opposite meaning. Finally, we converted each candidate to a sentence with its opposite meaning and reversed the label. The modified candidate is then added to the final test set. We used the following three methods to generate the sentences with opposite meanings.

\begin{enumerate}
    \item \textbf{Adding/removing negation expressions}: We add or remove Korean negation expressions (e.g., ``안'', ``못'', ``~지 않다'').
    \item \textbf{Antonym replacement}: A word is replaced with its antonym.
    \item \textbf{Using both method 1 and 2 or idiom}: Both methods described above are used. If a sentence includes an idiom, we replace it with its corresponding opposite meaning idiom.
\end{enumerate}

\paragraph{Guidelines} We instructed the annotators to comply with the following guidelines to generate data points for the KB-SentiNeg task.

\begin{enumerate}
    \item The sentence should not include the name of specific brands or products. This guideline is meant to avoid any possible legal issues.
    
    \item To generate a new sentence resembling a real product review, typos and spacing errors that frequently occur in Korean spoken language should be included occasionally.
\end{enumerate}

\subsection{Evaluation Metrics}
All of our downstream tasks have discrete labels. Therefore, we use the $F1$ score as a base criterion to evaluate models' performance.

\section{Design Principles}
In this section, we provide detailed descriptions about how we attempted to achieve the design principles illustrated in Section~\ref{section:intro}.

\subsection{Human-driven data annotation}
Automatic data generation using meta-information, such as a review score and news article category (e.g., Naver Sentiment Movie Corpus\footnote{\href{https://github.com/e9t/nsmc}{https://github.com/e9t/nsmc}}, is a widely used approach to rapidly collect a large amount of labelled data. While it is an efficient approach, there exists a high risk of the dataset containing incorrect and ambiguous data points. Such noisy data points are a major issue for evaluation datasets because they can lead to spurious performance increases (and or degradations) in the performance of \acp{LM}.

Our data is created purely by human annotations to produce the highest quality dataset with the lowest amount of incorrect and ambiguous data points possible. We hired four annotators who are Korean native speakers and major in \textit{Korean Language Education} or \textit{Korean Language in Literature}. Also, our Korean linguists trained the annotators before the data annotation process to avoid generating possible grammatical errors and unethical expressions.

\subsection{Leveraging linguistic knowledge}
Korean benchmark datasets, translated from English datasets (e.g., Kornli and Korsts~\cite{ham2020kornli}), might include incorrect translations and grammatical errors, particularly if they are machine-translated. Moreover, since the original examples come from English, such benchmark datasets are unlikely to assess properly assess Korean-specific knowledge or language intricacies.

Relying on our in-house Linguistic team, allowed us to mitigate and resolve issues with automatically generated datasets. First, our linguists trained the annotators to generate natural and grammatical Korean sentences and performed thorough reviews of the data. Thanks to their efforts, we have created a highly grammatical and natural. Secondly, the linguists designed tasks and data generation processes that considered the Korean language's characteristics. This is illustrated in guidelines for each task in Section~\ref{section:KoBEST}. Such guidelines and processes enabled us to create an accurate Korean evaluation dataset with expressive vocabulary and colloquial usage.

\subsection{Availability to public}
Our data is free from copyright issues. All sentences and answers, except for the paragraphs in KB-BoolQ task, are generated by our annotators from scratch by referencing publicly available sources. Also, the paragraphs in KB-BoolQ were extracted from Wikipedia, which is free from copyright issues. Therefore, researchers are free to use, modify and redistribute the KoBEST dataset.

\begin{table}[t]
	\begin{center}
		\renewcommand{\arraystretch}{1.1}
		\footnotesize{
			\centering{\setlength\tabcolsep{2pt}
		\begin{tabular}{l}
		\toprule
		\textbf{Example \#1} \\ \hline
		\ins{\textbf{P}: 날씨가 추워졌다. (The weather has become colder.)} \\
		\textbf{Q}: 원인 (Cause) \\
		\inp{\textbf{A1}: 겨울이 되었다. (Winter has come.)} \\
		\textbf{A2}: 여름이 되었다. (Summer has come.)\\ \hline
		
	    \textbf{Example \#2} \\ \hline
		\inp{\textbf{P}: 겨울이 되었다. (Winter has come.)} \\
		\textbf{Q}: 결과 (Effect) \\
		\ins{\textbf{A1}: 날씨가 추워졌다. (The weather has become colder.)} \\
		\textbf{A2}: 날씨가 더워졌다. (The weather became hot.)\\
		\bottomrule
		\end{tabular}}}
	\vspace*{-2ex}
	\caption{Examples of the KB-COPA data where the causality is interlocked. \textbf{P}, \textbf{Q}, \textbf{A1} and \textbf{A2} denote a premise, question, and alternatives, respectively. The two sentences highlighted in red and blue colours are swapped by changing the question from \textit{Cause} to \textit{Effect}.}\label{table.copa_wrong_examples}%
	\end{center}
	\vspace*{-4ex}
\end{table}

\begin{figure*}[t]
	\centering
	\begin{subfigure}[b]{1.0\textwidth}
		\includegraphics[width=\linewidth]{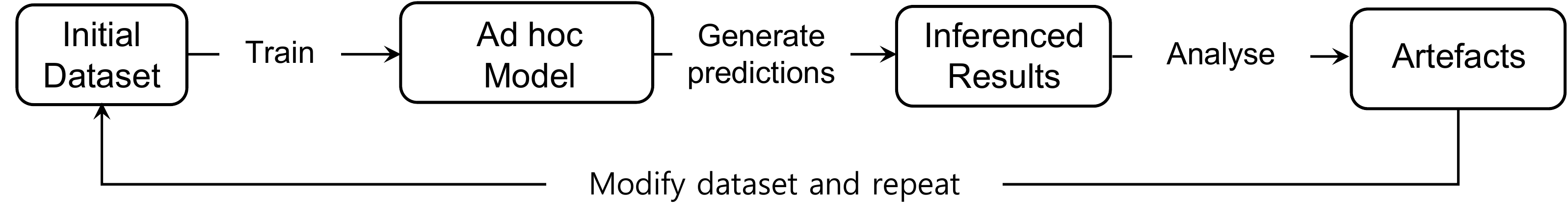}
		\vspace*{-2ex}
	\end{subfigure} \
	\vspace*{-2ex}
	\caption{The overall process of model-driven review for removing artefacts in training data.}
	\vspace*{-2ex}
	\label{fig:model_review_process}
\end{figure*}

\subsection{High data quality}\label{section:high_data_quality}
As a benchmark suite, accomplishing high data quality is important to accurately evaluate various \acp{LM}. We achieve this purpose through two review phases: \textit{human-driven} and \textit{model-driven} reviews.

\paragraph{Human-driven review} After collecting data, our linguists reviewed all data instances and found two major issues in the KB-COPA and KB-HellaSwag datasets.

For the KB-COPA task, we observed many cases with high correlation between between data instances. Examples corresponding to this case are presented in Table~\ref{table.copa_wrong_examples}. We conjecture this is because it was easier for the annotators to collect data by simply swapping the premise and answer along with changing the question. We removed or modified such instances because they are near duplicates and harm data diversity.

In the KB-HellaSwag task, we found several critical cases where predicting a correct final sentence is quite difficult due to the omission of detailed delineations in the context. This issue occurs because the source that the annotators referenced for generating contexts lacks detailed information occasionally. For such ambiguous instances, we appended additional clues to the context to allow inferring the meaning between the context and the final sentence.

\paragraph{Model-driven review} Artefacts existing in training data can lead a model to learn spurious inductive biases, resulting in distorted evaluation results~\cite{gururangan2018annotation, mccoy2019right, hossain2020analysis}. Therefore, we conduct a model-driven review process to find and remove such unwanted artefacts. The overall process is illustrated in Figure~\ref{fig:model_review_process}. First, we trained an ad-hoc model with the initial dataset for each task. Next, we generated predictions for the development and test datasets and analysed the results to ascertain whether specific words/patterns/phrases were strongly correlated with labels. Finally, our linguists analyzed the issues and updated the datasets accordingly. We repeated the whole process up to three times for each dataset. Through this process, we observed serious artefacts, especially in the KB-WiC and KB-COPA datasets. More than 70\% of questions containing number-related representations had \textit{False} as a label. Also, the label distributions of data instances containing specific phrases (e.g., "
덥다/춥다 (hot/cold)" or 했다/하지 않았다 (did/did not)) were highly skewed towards the \textit{False} label. All such artefacts were successfully removed and modified by our linguists.

\subsection{Avoiding AI ethical issues}
Social biases embedded in training data can lead to unethical behaviour of language models~\cite{nangia2020crows}. To mitigate such issues, we made efforts to remove unethical expressions, such as toxic content (e.g., insults, slang, sexual harassment) and social bias (e.g., gender, race, religion). Our linguists clearly instructed the annotators to avoid unethical expressions when generating sentences and extracting paragraphs from Wikipedia for the KB-BoolQ task. Also, linguists reviewed the data for potential ethical issues, as described in Section~\ref{section:high_data_quality}

\begin{table*}[t!]
	\begin{center}
		\renewcommand{\arraystretch}{1.2}
		\footnotesize{
			\centering{\setlength\tabcolsep{2.0pt}
		\begin{tabular}{c|c|c|c|c|c|c}
		\toprule
		\multicolumn{1}{c|}{\multirow{2}{*}{Model}} & \multicolumn{1}{c|}{KB-BoolQ} & \multicolumn{1}{c|}{KB-COPA} & \multicolumn{1}{c|}{KB-WiC} & \multicolumn{1}{c|}{KB-HellaSwag} & \multicolumn{1}{c|}{KB-SentiNeg} & 
		\multicolumn{1}{c}{Average} \\ 
		& $\mathcal{F}_{test}$ & $\mathcal{F}_{test}$ 
		& $\mathcal{F}_{test}$ & $\mathcal{F}_{test}$ & $\mathcal{F}_{test}$ &  $\mathcal{F}_{test}$\\ \hline
		
		KoBERT (FT) & 62.9$\pm$3.0 & 74.6$\pm$0.8 & 77.3$\pm$0.8 & 74.4$\pm$0.4 & 86.8$\pm$2.0 & 75.2 \\
		
		KoElectra (FT) & \textbf{75.1$\pm$1.0} & \textbf{81.5$\pm$0.4} & \textbf{79.7$\pm$1.8} & \textbf{74.7$\pm$0.8} & \textbf{91.9$\pm$1.1} & \textbf{80.6} \\
		
		KoGPT3-1.2B (FT) & 73.5$\pm$1.6 & 79.3$\pm$0.6 & 68.4$\pm$2.2 & 73.8$\pm$1.0 & 89.5$\pm$3.3 & 77.0 \\
		
        KoBART (FT) &60.6$\pm$2.9 & 56.9$\pm$3.2 &  60.4$\pm$4.9 & 51.4$\pm$1.3 & 88.6$\pm$1.2 & 63.6 \\ \hline
        
        KoGPT3-39B ($k=0$) & 33.1 & 76.8 & 34.7 & 59.8 & 57.7 & 52.8 \\
        KoGPT3-39B ($k=1$) & 50.2 & 78.3 & 51.8 & 60.2 & 74.2 & 66.3 \\
        KoGPT3-39B ($k=10$) & 46.9 & 80.9 & 52.2 & 58.7 & 91.6 & 70.6 \\ \hline
        Human & 95.1 & 98.1 & 96.6 & 92.4 & 99.0 & 96.2 \\
		\bottomrule
		\end{tabular}}}
	\end{center}
	\caption{The performance of Korean \acp{LM} on the KoBEST downstream tasks. $\mathcal{F}_{test}$ refer to the test F1 score. The first and second blocks are the experimental results of fine-tuning and zero/few-shot learning, respectively. For the fine-tuning experiments, we repeat each experiment five times and report the average and standard deviation. The best values for each task are written in bold.}\label{table.finetuning_result}
\end{table*}

\section{Experiments}
In this section, we provide model and human performance baseline results.

\subsection{Fine-tuning Experiments}
\subsubsection{Experimental Design}
\paragraph{Model Candidates} We used the following four pre-trained Korean language models to benchmark our KoBEST dataset:

\begin{itemize}
    \item \textbf{Encoder models}: KoBERT~\footnote{\href{https://huggingface.co/monologg/kobert}{https://huggingface.co/monologg/kobert}}, KoElectra~\footnote{\href{https://huggingface.co/monologg/koelectra-base-v2-discriminator}{https://huggingface.co/monologg/koelectra-base-v2-discriminator}}
    \item \textbf{Decoder models}: KoGPT3-1.2B~\footnote{\href{https://huggingface.co/skt/ko-gpt-trinity-1.2B-v0.5}{https://huggingface.co/skt/ko-gpt-trinity-1.2B-v0.5}}
    \item \textbf{Encoder-Decoder models}: KoBART~\footnote{\href{https://huggingface.co/hyunwoongko/kobart}{https://huggingface.co/hyunwoongko/kobart}}
\end{itemize}

For KoBART, we applied the text-to-text multi-task training technique to train the model. We transformed inputs of each text to a free-text input by referencing the transformation formats used in the T5 model~\cite{T5}.

\begin{table}[t]
	\begin{center}
		\renewcommand{\arraystretch}{1.0}
		\footnotesize{
			\centering{\setlength\tabcolsep{2.0pt}
		\begin{tabular}{c|c|c|c|c|c}
		\toprule
		\multirow{2}{*}{} & 
		BoolQ & COPA & WiC & HellaSwag & SentiNeg \\ \hline
		b-size & 8 & 16 & 16 & 8 & 16 \\
		s-len & 256 & 128 & 256 & 256 & 128  \\
		lr & $5e^{-6}$ & $5e^{-6}$ & $1e^{-5}$ & $2e^{-5}$ & $1e^{-5}$ \\
		\bottomrule
		\end{tabular}}}
	\vspace{-1ex}
	\caption{Batch-size (b-size), maximum input length (s-len), and learning rates (lr) used for the KoBEST benchmark experiments.}
	\vspace{-1ex}\label{table.hyperparam}%
	\end{center}
	\vspace{-1ex}
\end{table}

\paragraph{Training Details} We used AdamW optimiser~\cite{loshchilov2018fixing} for training with a linear learning rate scheduler decaying from 1\textit{e}-2. We trained all models for 10 epochs, and used the early stopping method during the training. Different batch size and learning rate were used across the tasks, and detailed training hyperparamters are presented in  Table~\ref{table.hyperparam}.

\subsubsection{Results and Discussion} The fine-tuning results can be seen in the first block of Table~\ref{table.finetuning_result}. Overall, KoElectra showed the best results, followed by KoGPT-1B, suggesting that model size is not necessarily a requisite for better performance. 

\begin{table}[t!]
	\begin{center}
		\renewcommand{\arraystretch}{1.2}
		\footnotesize{
			\centering{\setlength\tabcolsep{2.0pt}
		\begin{tabular}{c|c|c|c}
		\toprule
		\multicolumn{1}{c|}{Model} & \multicolumn{1}{c|}{$\mathcal{F}_{val}$} & \multicolumn{1}{c|}{$\mathcal{F}_{test}$} & \multicolumn{1}{c}{$\triangle$} \\ \hline
		
		KoBERT & 99.1 & 86.8 & 12.3 \\
		KoElectra & 99.4 & 91.9 & 7.5 \\
		KoGPT3-1B & 99.8 & 89.5 & 10.3 \\
		KoBART & 98.7 & 88.6 & 10.1 \\
		\bottomrule
		\end{tabular}}}
	\end{center}
	\caption{The average performance gap between validation and test sets of KB-SentiNeg task.}\label{table.sentineg_diff}
\end{table}

\paragraph{Do models understand the opposite in meaning?}  Table~\ref{table.sentineg_diff} shows the validation and test performance of fine-tuned models on the \textbf{SentiNeg} task. Interestingly, all models show a large performance gap between the validation and test performance in the \textbf{SentiNeg} task. The results suggest that \acp{PLM} are vulnerable to a simple negation and antonym-replacement perturbation, even though the data points all originated from the training data. Our results are aligned with previous studies on English data that showed \acp{PLM} are incapable of understanding negation expressions~\cite{hossain2020analysis, kassner2020negated, ettinger2020bert, hosseini2021understanding}, suggesting that the issue stems from the \ac{PLM}, not from language itself.

\paragraph{Multi-task training is not always beneficial}
Unlike the other three models, KoBART was trained in a multi-task fashion. However, contrary to the common belief that multi-task training is beneficial in improving performance on benchmark suites (e.g., GLUE~\cite{mtdnn}), in our case, multi-task training produced the worst performance by a large margin. We conjecture that a leading cause is a misalignment between tasks. All the downstream tasks in KoBEST are independent of each other. However, in the GLUE benchmark, for instance, the sub-tasks are well aligned, containing multiple datasets that share a common objective, e.g., \ac{NLI} and \ac{STS}, and it is well studied that the misalignment between task data can cause poor results~\cite{wu2020understanding}.

\subsection{Zero/Few Shot Experiments}
\subsubsection{Experimental Design}
The advent of extremely large size \acp{GLM} like GPT3~\cite{GPT3} has allowed in-context learning (providing the model with a few or no samples) to apply the model to downstream tasks. To this end, we conducted zero-, one- and few-shot experiments by using a Korean GPT3 model Language Superintelligence Labs trained with 39 billion parameters and trained with 132 billion tokens. We then referenced the work of EleutherAI~\footnote{\href{https://github.com/EleutherAI/lm-evaluation-harness}{https://github.com/EleutherAI/lm-evaluation-harness}} to design prompts for our zero, one, and few-shot experiments. Several prompt examples are available in Table~\ref{table.prompt_design} in the appendix.  For multiple-choice problems like \textbf{KB-COPA} and \textbf{KB-HellaSwag}, we selected the candidate having the lowest perplexity as the prediction.

\subsubsection{Results and Discussion}
\paragraph{Fine-tuned models are still best.} The results, presented in the second block of Table~\ref{table.finetuning_result}, reveal that the fine-tuned models, apart from KoBART, outperform in-context learning methods. Our results are aligned with the work of \citet{GPT3} that showed GPT3 performance based on few-shot learning is behind that of fine-tuned SOTA in many tasks, including all downstream tasks of SuperGLUE~\cite{SuperGLUE}. Although it is interesting that a large \ac{GLM} can achieve decent performance with only a few training examples, results suggest that we should be judicious using large \ac{GLM} in practical applications; especially when considering performance compared to excessively high training costs (i.e., time and resources).

\paragraph{Increasing $k$ is not always beneficial.} Few-shot learning approaches with more examples increase performance in general, but merely increasing $k$ does not always lead to better performance. Specifically, in the case of KoBEST, the performance is slightly worse on the \textbf{KB-BoolQ} and \textbf{KB-HellaSwag} tasks. We believe that the length of the input document is a leading cause of this phenomenon. Since the model's max input length plays a critical role in deciding the maximum number of examples ($n$) in the prompt~\cite{yang2021empirical}, the available $n$ decreases as the length of prompts increases. However, as we can see in Table~\ref{table.examples}, the data points of the \textbf{KB-BoolQ} and \textbf{KB-HellaSwag} tasks have longer inputs than the other tasks. As a result, the prompts for these tasks become very long and likely to exceed the model's maximum input length. This would result in a sliced prompt that may lack key information the model needs to make a successful prediction.

\subsection{Human performance}
We asked volunteers to evaluate the dataset to provide human-level performance metrics for KoBEST. Specifically, 10 native Korean evaluators evaluated 100 randomly sampled examples for each downstream task. The results are summarised in the last row of Table~\ref{table.finetuning_result}. The human evaluators outperformed all the \acp{PLM} by a large margin, suggesting that modern \acp{PLM} need further improvements to achieve human-level language ability.

\section{Conclusion}
A well-designed benchmark dataset is crucial for an objective and precise evaluation of \acp{LM}. Following the GLUE benchmark, more challenging benchmarks have been proposed as modern \acp{LM} become more elaborate and sophisticated. However, most of these benchmarks only support English or originate from English (e.g., translation), which hardly captures important characteristics of a specific language.

To this end, we propose a new Korean benchmark suite named \ac{KoBEST}, which consists of five challenging downstream tasks. To overcome the disadvantages of the previous Korean benchmarks, we focused on 1) employing Korean-specific knowledge, 2) achieving high data quality and 3) removing superficial cues. To achieve these goals, we worked with professional Korean linguists and collected data manually and not automatically. We also conducted human- and model-driven review processes to eliminate superficial cues from our dataset. Moreover, we were extra cautious to avoid using unethical expressions.

Finally, we evaluated various \acp{PLM} on our new benchmark and provide baseline model and human performance metrics. Our experimental results show that current \acp{LM} need further improvements to attain human-level language ability. We hope our new benchmark can contribute to advancements in the Korean \ac{NLP} field.

\begin{acronym}
    \acro{LM}{language model}
    \acro{MLM}{masked language modelling}
    \acro{NLI}{natural language inference}
    \acro{NLP}{natural language processing}
    \acro{NLU}{natural language understanding}
    \acro{STS}{semantic textual similarity}
    \acro{NER}{named entity recognition}
    \acro{RE}{relation extraction}
    \acro{GLM}{generative language model}
    \acro{PLM}{pre-trained language model}
    \acro{KoBEST}{\textbf{Ko}rean \textbf{b}alanced \textbf{e}valuation of \textbf{s}ignificant \textbf{t}asks}
    \acro{BoolQ}{boolean questions}
    \acro{COPA}{choice of plausible alternatives}
    \acro{WiC}{words in context}
    \acro{QA}{question answering}
    \acro{PoS}{part of speech}
    \acro{KLUE}{Korean Language Understanding Evaluation}
\end{acronym}

\section*{Acknowledgements}
We would like to express our great appreciation to Sunwoo Lee, Seo-jin Lee, and Seokyoung Hong for their valuable assistance and feedback.

\bibliography{reference}
\bibliographystyle{acl_natbib}

\clearpage
\appendix

\section{Appendix}
\label{sec:appendix}

\begin{table*}[ht]
	\begin{center}
		\renewcommand{\arraystretch}{1.2}
		\footnotesize{
			\centering{\setlength\tabcolsep{1.5pt}
		\begin{tabular}{c|l}
		\toprule
		\rotatebox[origin=c]{90}{\textbf{KB-BoolQ}} & \makecell[l]{
		\textbf{Inputs}: \\
		Paragraph: 구한말, 통영 안뒤산 기슭 간창골에 김봉제 형제가 살았다. 김봉제는 관약국을 경영하며 부를 누렸는데, \\ 선비적 성품을 지녔던 형과 반대로 막냇동생 김봉룡은 성질이 포악했다. 어느 날 봉룡은 아내였던 숙정을 사모하던 \\ 나그네를 살해하였고, 숙정은 누명을 벗으려고 비상을 먹고 자살한다. \\ Question: 봉룡은 숙정을 죽였는가?  \quad Answer: False \\
		\textbf{Prmopt Design}: \\
		Answer: ``예'' if Answer is True else ``아니오'' \\
		Format: ``\{Paragraph\} 질문: \{Question\} 답변: \{Answer\}.'' \\
		\textbf{Example}: \\
		구한말, 통영 안뒤산 기슭 간창골에 김봉제 형제가 살았다. 김봉제는 관약국을 경영하며 부를 누렸는데, \\ 선비적 성품을 지녔던 형과 반대로 막냇동생 김봉룡은 성질이 포악했다. 어느 날 봉룡은 아내였던 숙정을 사모하던 \\ 나그네를 살해하였고, 숙정은 누명을 벗으려고 비상을 먹고 자살한다. 질문: 봉룡은 숙정을 죽였는가? 답변: 아니요.
		} \\ \hline
		
		\rotatebox[origin=c]{90}{\textbf{KB-COPA}} & \makecell[l]{
		\textbf{Inputs}: \\
		Premise: 전쟁이 시작되었다. \quad Question: 결과 \\ Answer Alternative: 병사들이 전투에 파견되었다. \\
		\textbf{Prompt Design}: \\
		Connector: ``왜냐하면'' if Question is ``원인'' else ``그래서'' \\
		Format: ``\{Premise\} \{Connector\} \{Answer Alternative\}'' \\
	    \textbf{Example}: \\
		전쟁이 시작되었다. 그래서 (``왜냐하면'' if question is 원인) 병사들이 전투에 파견되었다.} \\ \hline
		
		\rotatebox[origin=c]{90}{\textbf{KB-WiC}} & \makecell[l]{
		\textbf{Inputs}: \\
		Context 1: 망가진 엔진은 수리가 불가능하다.\\ Context 2: 이 배는 수리에 들어간 지 일주일이 됐다. \\ Target Word: 수리 \quad Answer: True \\
		\textbf{Prompt Design}: \\
		Answer: ``예'' if Answer is True else ``아니오'' \\
		Format: ``문장1: \{Context1\} 문장2: \{Context2\} 두 문장에서 \{Target Word\}가 같은 뜻으로 쓰였나? \{Answer\}'' \\
		\textbf{Example}: \\
	    문장1:  망가진 엔진은 수리가 불가능하다. 문장2: 이 배는 수리에 들어간 지 일주일이 됐다. 두 문장에서 수리가 같은 \\ 뜻으로 쓰였나? 예.} \\ \hline
		
		\rotatebox[origin=c]{90}{\textbf{KB-HellaSwag}} & \makecell[l]{
		\textbf{Inputs}: \\
		Context: 양궁 선수들이 경기장으로 입장한다. 관중들이 함성을 지르고 응원한다. 선수들이 상대팀과 악수하고 \\ 자리로 돌아온다. 코치가 전략을 설명하고 화이팅을 외친다. \\ Correct Ending: 선수들이 각자 자리에 서서 활을 꺼낸다.  \\ 
		\textbf{Prompt Design}: \\
		문장: \{Context\} \{Correct Ending\} \\
		\textbf{Example}: \\
		문장: 양궁 선수들이 경기장으로 입장한다. 관중들이 함성을 지르고 응원한다. 선수들이 상대팀과 악수하고 \\ 자리로 돌아온다. 코치가 전략을 설명하고 화이팅을 외친다. 선수들이 각자 자리에 서서 활을 꺼낸다.
		} \\ \hline
		
		\rotatebox[origin=c]{90}{\textbf{KB-SentiNeg}} & \makecell[l]{
		\textbf{Inputs}: \\
		Sentence: 뚜껑이 잘 안열려요! \quad Answer: 부정 \\ 
		\textbf{Prompt Design}: \\
		Format: ``문장: \{Sentence\} 긍부정: \{Answer\}'' \\
		\textbf{Example}: \\
		문장: 뚜껑이 잘 안열려요! 긍부정: 부정 \\
		} \\
		\bottomrule
		\end{tabular}}}
	\vspace*{-2ex}
	\caption{Prompt designs for each task used in in-context learning experiments. Example data points presented in Table~\ref{table.examples} are used.}\label{table.prompt_design}%
	\end{center}
	\vspace*{-4ex}
\end{table*}

\end{document}